\documentclass[10pt,twocolumn,letterpaper]{article}
\usepackage[accsupp]{axessibility}
\usepackage{iccv}
\usepackage{times}
\usepackage{epsfig}
\usepackage{graphicx}
\usepackage{amsmath}
\usepackage{amssymb}

\usepackage{algorithm,algorithmic}

\usepackage{verbatim}
\usepackage{graphicx} %
\usepackage{subfloat}
\usepackage{tikz}
\usepackage{pgfplots}

\DeclareMathOperator*{\Sign}{Sign} 
\DeclareMathOperator*{\XNOR}{XNOR} 
\DeclareMathOperator*{\NOTs}{NOT}
\DeclareMathOperator*{\Popcount}{Popcount}

\DeclareMathOperator*{\argmin}{argmin} 
\DeclareMathOperator*{\inputs}{input}
\DeclareMathOperator*{\w}{w}
\usepgfplotslibrary{groupplots}
\usepackage{multirow}
\usepackage[flushleft]{threeparttable}

\usepackage[pagebackref=true,breaklinks=true,letterpaper=true,colorlinks,bookmarks=false]{hyperref}

\iccvfinalcopy 

\def\iccvPaperID{11850} 

\ificcvfinal\fi

\begin{document}

\title{MST-compression: Compressing and Accelerating Binary Neural Networks \\with Minimum Spanning Tree}

\author{Quang Hieu Vo, Linh-Tam Tran, Sung-Ho Bae, Lok-Won Kim, Choong Seon Hong\thanks{Corresponding Author}\\
Department of Computer Science and Engineering, Kyung Hee University \\
{\tt\small \{2019310178, tamlt, shbae, lwk, cshong\}@khu.ac.kr}
}

\maketitle

\ificcvfinal\thispagestyle{empty}\fi

\begin{abstract}
Binary neural networks (BNNs) have been widely adopted to reduce the computational cost and memory storage on edge-computing devices by using one-bit representation for activations and weights.
However, as neural networks become wider/deeper to improve accuracy and meet practical requirements, the computational burden remains a significant challenge even on the binary version.
To address these issues, this paper proposes a novel method called Minimum Spanning Tree (MST) compression that learns to compress and accelerate BNNs.
The proposed architecture leverages an observation from previous works that an output channel in a binary convolution can be computed using another output channel and $\XNOR$ operations with weights that differ from the weights of the reused channel.
We first construct a fully connected graph with vertices corresponding to output channels, where the distance between two vertices is the number of different values between the weight sets used for these outputs.
Then, the MST of the graph with the minimum depth is proposed to reorder output calculations, aiming to reduce computational cost and latency.
Moreover, we propose a new learning algorithm to reduce the total MST distance during training. Experimental results on benchmark models demonstrate that our method achieves significant compression ratios with negligible accuracy drops, making it a promising approach for resource-constrained edge-computing devices.

\end{abstract}

\section{Introduction}
\begin{figure}[t]
\begin{center}
   \includegraphics[width=0.6\linewidth]{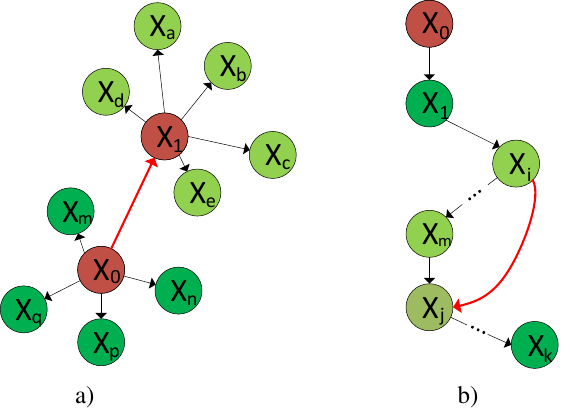}
\end{center}
\caption{Illustration of kernel compression for BNNs with the K-mean method (a) and the shortest Hamiltonian path (b), in which adding red connections can further reduce the computational cost for both of them.
}
\vspace{-0.4cm}
\label{fig:k-mean_halmi}
\end{figure}
Deep Neural Networks (DNNs) have been widely applied in many artificial intelligence applications, especially vision tasks with high accuracy \cite{sandler2018mobilenetv2,he2016deep}.
However, the cost of computation and massive storage burden is significantly challenging to deploy DNNs on embedded systems such as mobile devices and other resource-constrained platforms.
Many approaches have been proposed and demonstrated their effectiveness in reducing energy and resources while maintaining the deep models' accuracy, including pruning \cite{han2015deep}, quantization \cite{courbariaux2016binarized}, distillation \cite{hinton2015distilling}, and efficient hardware implementation \cite{chen2016eyeriss}.
Among these methods, quantization with less bit-width for parameter and activation representation is widely used due to its great benefits and possibility in most practical applications.

Binarized neural network is a particular form of the quantization method in which weight values and activations are converted to 1-bit values.
Accordingly, multiplications and accumulation operations can be replaced by $\XNOR$ and $\Popcount$ operations \cite{courbariaux2016binarized}, respectively.
In addition, batch normalization is simplified to a threshold comparison \cite{vo2022deep}, while the pooling can be performed with OR operations \cite{umuroglu2017finn}.
Consequently, the compression ratio on memory storage and computational cost is significantly improved, leading to remarkable performance acceleration.
Nevertheless, the accuracy degradation is the trade-off of minimizing the bit-width as BNN.
Thus, most previous works focus on reducing this accuracy gap \cite{rastegari2016xnor, bulat2019xnor,xu2021recu,liu2020reactnet,lin2020rotated, xu2021recu, liu2018bi}.

\begin{figure}[t]
\begin{center}
   \includegraphics[width=1\linewidth]{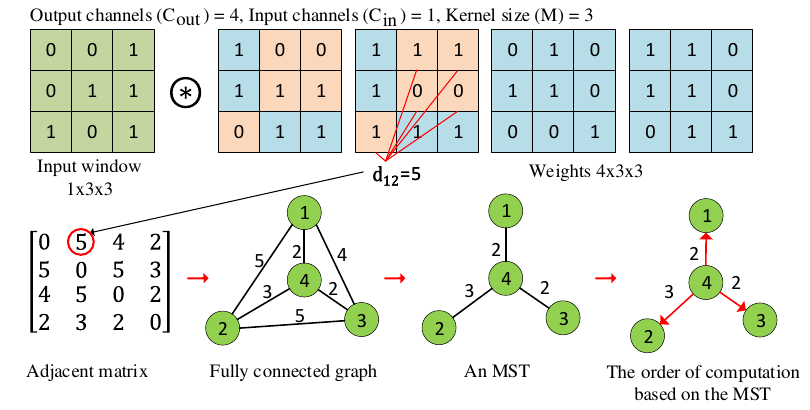}
\end{center}
\caption{The process of arranging the order of computation on a binary convolution layer, in which output channel 4 is fully calculated first. Then, the output channel 4 is used to calculate channel 1, channel 2 and channel 3.}
\label{fig:MST_diagram}
\end{figure}

Meanwhile, compressing BNNs has not received much attention, with only a few prominent methods \cite{wang2021snn, fu2018towards, vo2022deep, kim2017kernel}, where the kernel compression \cite{fu2018towards, vo2022deep} gives impressive results with roughly $50$\% resources reduction.
In particular, given a binary convolution, inspired by an observation that an output channel can be calculated using another output channel and outputs of $\XNOR$ operations using weights that respectively differ from the reused channel's weights \cite{fu2018towards}, authors in \cite{vo2022deep, fu2018towards} proposed to construct a fully connected graph, in which each vertex corresponds to an output channel, and distance between two vertices is the number of different values between two weight sets used for the two output channels.
Then, based on the graph, they use the K-mean cluster and shortest Hamiltonian path to reorder the convolution output calculation, aiming to reduce the number of $\XNOR$ operations.
However, these approaches have yet to fully minimize the computational cost as output channels can use less computation.
Indeed, Figure \ref{fig:k-mean_halmi} (a) illustrates an example of the K-mean method, where two vertices are considered centers of two groups.
Two output channels corresponding to centers are fully calculated\footnote{calculated with $C_{in}\times M\times M \XNOR$s, denoted in Figure \ref{fig:MST_diagram}}, while other vertices reuse their centers for calculation.
However, adding a connection between the two centers would enable one to utilize the rest with less computation cost.
Additionally, in the shortest Hamiltonian path shown in Figure \ref{fig:k-mean_halmi} (b), there may exist a connection to a specific vertex that is shorter than the connection from the preceding vertex in the path, leading to a lower required number of operations for this vertex.
Furthermore, time complexity poses a challenge in these methods \cite{bellman1962dynamic, hartigan1979algorithm}, resulting in longer exploration times.

To more effectively minimize the number of $ \XNOR $ operations with the mentioned observation, in all connections to a specific vertex, the minimum connection must be selected to minimize the computation cost for this vertex.
In addition, all of these selected connections must be included in a subgraph that visits all vertices exactly once to ensure that only one output channel is fully calculated.
As a result, a MST is the only structure that can fulfill these requirements. 
Therefore, this paper leverages the MST to reorder binary convolution output calculations.
Figure \ref{fig:MST_diagram} shows a simple example of reordering calculation on a convolution.
In this example, only the output channel 4 is fully calculated with $C_{in}\times M\times M$ bit-weight values.
In contrast, other channels are calculated using output channel $4$, following the MST direction.

On the other hand, to maximize the advantages of the MST, we further minimize the MST distance of all convolution layers with a new learning algorithm right during the training stage.
Besides, we propose a hardware accelerator for BNNs applying the MST compression to demonstrate the feasibility and effectiveness of the method related to hardware resources. 
To the best of our knowledge, this method gives the highest compression ratio with implementation from learning for compression to acceleration.
In summary, the following are contributions of this paper:

\begin{itemize}
    \item We introduce and analyze the effectiveness of the MST in reducing the computational cost for the inference on a binary convolution layer.
    \item We propose a training algorithm that can reduce the MST distance and depth right during the training process, which consequently maximizes the compression ratio for inference implementation.
    \item We provide the corresponding hardware acceleration for the proposed method with high throughput and better resource efficiency, compared to related works \cite{blott2018finn, umuroglu2017finn, vo2022deep, ghasemzadeh2018rebnet}.
\end{itemize}

The experiments are performed for BNNs including VGG-small \cite{cai2017deep}, ResNet-18/20 \cite{he2016deep} on CIFAR-10 dataset \cite{krizhevsky2009learning}, and ResNet-18/34 \cite{he2016deep} on ImageNet \cite{russakovsky2015imagenet}.
The results show that the proposed approach gives a higher compression ratio than the previous works \cite{wang2021snn}.
Compared to the baseline \cite{xu2021recu}, our method reduces up to $13.5\times$ the convolution parameters and $5.51\times$ bit-wise operations on the same model with an acceptable accuracy degradation.
Regarding hardware acceleration, we conduct the experiments for a BNN with the same structure as the baseline \cite{vo2022deep} and apply the proposed approach.
Compared to the baseline \cite{vo2022deep}, hardware deployments demonstrate that BNNs with our method save $1.8\times$ LUTs, and $1.81\times$ area efficiency while maintaining the acceleration speed and accuracy.

\section{Related Work}
Training neural networks with binary weights and activations were first introduced by Coubariaux \etal \cite{courbariaux2016binarized} and then rapidly gained attention from the community due to the high compression ratio.
However, accuracy drop is a critical problem of this direction, while compression is still highly demanded.
Various techniques are proposed to further compress and improve accuracy based on the binary property.

For accuracy improvement, Rastegari \etal \cite{rastegari2016xnor} proposed adding a scaling factor to the convolution to reduce the quantization error.
To generalize this idea aiming to enhance accuracy, the authors in \cite{bulat2019xnor} extended this factor dimension and enabled its training option as a learnable parameter.  
Besides, instead of using the Straight Through Estimation (STE) method \cite{bengio2013estimating}, Liu \etal \cite{liu2018bi}, and Lin \etal \cite{lin2020rotated} proposed using a piece-wise linear and training-aware approximation function, respectively, to approximate the sign function gradient better.
Meanwhile, some previous works introduced both new optimization techniques and corresponding BNN to shrink the accuracy gap entirely \cite{liu2020reactnet, lin2017towards, tu2022adabin, qin2020forward}.
For example, based on Bi-real Net in \cite{liu2018bi}, authors in \cite{liu2020reactnet} proposed new sign and activation functions to shift and reshape activation distribution with a new baseline neural network model that can boost the performance.
Other works look at the binary characteristics from deeper perspectives to limit information loss \cite{lin2020rotated, tu2022adabin, xu2021recu}.
For example, KL divergence is employed in \cite{tu2022adabin} to minimize the difference between the distribution of binary and real-weight counterparts.
Xu \etal \cite{xu2021recu} observed the low probability of changing large full-precision weight when binarizing and proposed the ReCu function to revive the death weights, leading to lower quantization error.

Binary model compression has received less attention as only a few techniques have been proposed so far.
Specifically, Lee \etal \cite{lee2020flexor} proposed an encryption algorithm to compress binary weights with lower than 1-bit per weight, and XOR-gate networks are used to decrypt the inference task.
Similarly, Sub-bit Neural Networks (SNN) \cite{wang2021snn} proposed using lower than $9$ bits for each $3\times 3$ binary kernel based on the scattered $9$-bit kernel distribution.
Other proposed compression methods provided only hardware techniques for inference implementation.
Authors in \cite{vo2022deep,fu2018towards} proposed BNN hardware architectures using weight/input reuse with different calculation orders for convolution to reduce hardware overhead.
Kim \etal \cite{kim2017kernel} introduces kernel decomposition that can reduce the computational cost by sharing one base output for all channels.

\section{Background}
This section briefly describes how to construct a binary convolution layer and the related optimization methods.
For a convolution layer with $C_{in}$ input channels, $C_{out}$ output channels, and kernel size $M$, weights and activations are denoted by $\mathcal{W}^r$ $\in$ $\mathbb{R}^n$ and $A^r$ $\in$ $\mathbb{R}^m$, where $n = C_{out}\times C_{in}\times M\times M$ and $m = C_{in}\times W_{in}\times H_{in}$. 
$W_{in}\times H_{in}$ represents input feature map size. 
To simplify the computation and reduce memory storage, in a binary convolution layer, the binarization of activations $\mathcal{A}^b$ $\in$ $\{\pm1\}^m$ and weights $\mathcal{W}^b$ $\in$ $\{\pm1\}^n$ is acquired by the following sign function \cite{rastegari2016xnor},
\begin{equation} 
    x^b=\Sign(x)=\begin{cases}
                  +1,  \text{ if } x \geq 0,\\
                  -1,  \text{ otherwise.}
                 \end{cases}
\end{equation}
If $-1$ is substituted by $0$ to represent a negative value for $\mathcal{A}^b$ and $\mathcal{W}^b$, the output convolution operation of the $i^{th}$ channel now is reformulated as in Eq. (\ref{eq:weight-reuse}) \cite{vo2022deep}.
\begin{equation}
\label{eq:weight-reuse}
Y_i = (2\sum_{j=1}^{C_{in}MM}\XNOR(\mathcal{A}^b_{ij},\mathcal{W}^b_{ij}) - C_{in}\times M\times M) \odot \alpha,
\end{equation}
where $\sum_1^{C_{in}\times M\times M}\XNOR(\mathcal{A}^b_{ij},\mathcal{W}^b_{ij})$ is the number of set-bits ($\Popcount$) of the output channel $i^{th}$, $\odot$ is the element-wise multiplication, and $C_{in}\times M\times M$ is the number of bit-wise $\XNOR$ operations used to calculate one output channel.
The scaling factor $\alpha$ is added as a learnable parameter to lessen quantization error \cite{rastegari2016xnor}.
For backward propagation, following \cite{liu2018bi}, the STE \cite{bengio2013estimating} method is used for the approximation gradient on weights, while the piece-wise polynomial function \cite{liu2018bi} is used for gradient approximation on activations.
\begin{figure}[t]
\begin{center}
   \includegraphics[width=1\linewidth]{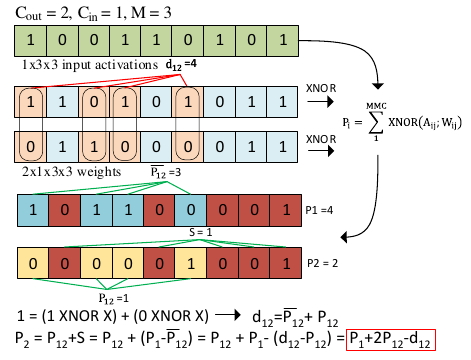}
\end{center}
\caption{Weight-reuse method description. Here, the red line is counting the number of elements, the green line is summing all elements. $P_{12}$ is $\sum_{j = 1}^{d_{12}}\XNOR (\mathcal{A}_{2j},\mathcal{W}_{2j})$, and $Y_2$ is $2(P_1-d_{12}+2P_{12})- C_{in}\times M\times M$.}
\vspace{-0.4cm}
\label{fig:weight-reuse}
\end{figure}

To further compress the BNN models, in this paper, we apply the weight reuse method \cite{vo2022deep} for the inference task.
Accordingly, given the $l^{th}$ binary convolution, as shown in Figure \ref{fig:weight-reuse}, we denote $P_i$ and $P_j$ as the output of the $\Popcount$ operations of the output channels $i^{th}$ and $j^{th}$, respectively.
All binary weights of the layer are divided into $C_{out}$ weight sets $w_b^{li} \in \{\pm1\}^{C_{in}\times M \times M}$, where $i = 1, 2, ..., C_{out}$ and each weight set is used to calculate for an output channel.
$d_{ij}$ is the number of weight bits in the $w_b^{li}$ that differ from the $w_b^{lj}$, compared one-to-one respectively.
$P_{ij}$ is the sum of $\XNOR$ operations with the weights of the channel $j^{th}$ that differ from the $i^{th}$ counterpart. 
To calculate the output channel $j^{th}$ from the output channel $i^{th}$, we use Eq. (\ref{eq:outputchannel}).
\begin{equation}
    \label{eq:outputchannel}
		Y_j = 2(P_i - d_{ij} + 2P_{ij})-C_{in}\times M\times M .
\end{equation}

\section{Methodology} \label{sec:methodology}
\subsection{Minimum Spanning Tree for BNN Acceleration} \label{sec:sub-MST_inference}
\begin{figure}[t]
\begin{center}
   \includegraphics[width=1\linewidth]{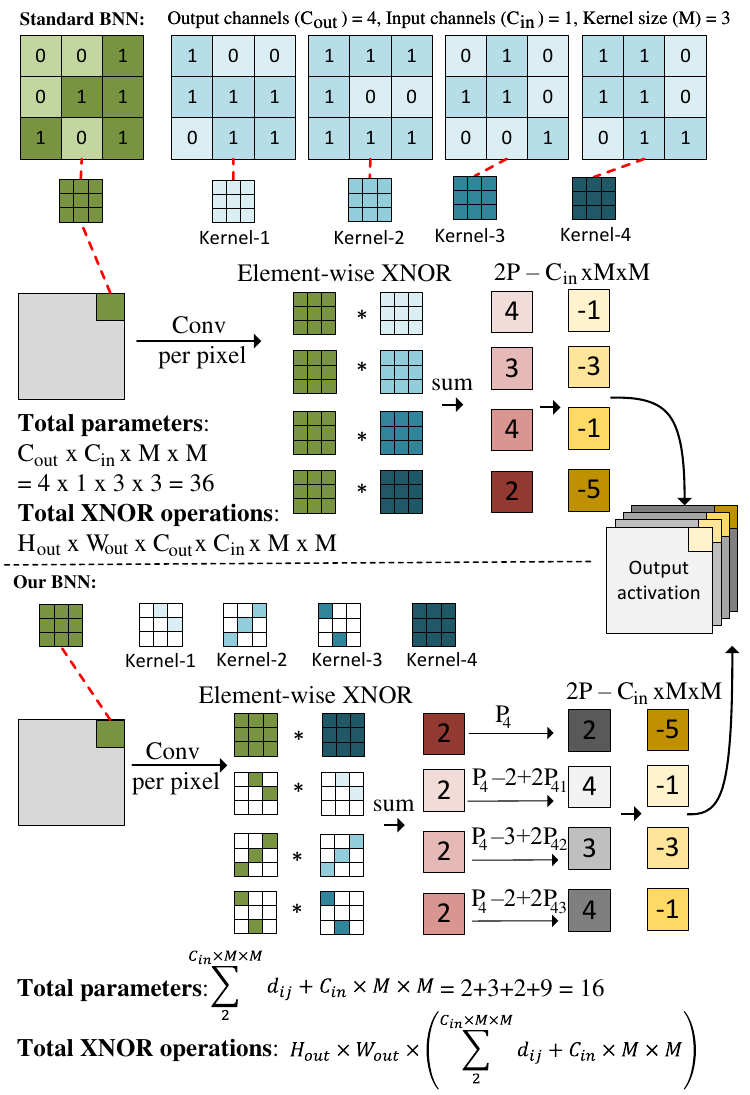}
\end{center}
\caption{Comparison between the computation process of a standard binary convolution and our method in BNN acceleration. Instead of fully computing all $C_{in}\times M \times M$ $\XNOR$ operations for each layer, our method provides an optimal order of the calculation using the MST to minimize the number of $\XNOR$ operations and total parameters.}
\vspace{-0.55cm}
\label{fig:computation-acceleration}
\end{figure}
For a binary convolution layer, we construct a fully connected graph in which each vertex corresponds to an output channel.
The distance between two vertices $i$ and $j$ is $d_{ij}$.
As shown in Eq. (\ref{eq:outputchannel}), an output channel $j$ can be calculated via the $\Popcount$ of an output channel $i$ and the $\Popcount$ of $d_{ij}$ $\XNOR$ operations. 
That means the number of $\XNOR$ operations executed for the output channel $j$ is the distance $d_{ij}$.
Based on this observation, to minimize the number of $\XNOR$ operations used for all output channels, we need a sub-graph that includes all vertices with the minimum total edge distances, which is named an MST \cite{tutte2001graph}.
Returning to the example in Figure \ref{fig:MST_diagram}, Figure \ref{fig:computation-acceleration} illustrates the computation process of this convolution example with the standard and proposed approach.
Specifically, in the standard direction, each output channel needs $9 \XNOR$ operations, while in our direction, only the $4^{th}$ output channel is fully computed by $9 \XNOR$ operations, and the $1^{st}$, $2^{nd}$, $3^{rd}$ output channels reuse the output channel $4^{th}$ with $2$, $3$, $2$ $\XNOR$ operations, respectively.
Accordingly, the compression ratio for this example is ($2+3+2+3\times 3$)/($4\times 1\times 3\times 3) = 0.44$.
In general, we can reduce the computational cost for a convolution layer with the following ratio:
\begin{equation}
     \label{eq:ratio-param}
    \mathcal{R} = \frac{\sum_{j\ne root}^{C_{out}}d_{ij} + C_{in} \times M \times M}{C_{out} \times C_{in} \times M \times M},
\end{equation}
where $i$ is the parent of the vertex $j$.
If $j$ is the root of the MST, this output uses $C_{in}\times M \times M$ binary weight values.

Regarding parameters, because the streaming architecture described in Sec. \ref{sec:hw_architecture} is fully implemented with all layers on the FPGA platform, weights' locations can be converted to connections from inputs to the correct $\XNOR$s and corresponding weight values at the circuit level.
Thus, our proposed method counts only the parameters we need for $\XNOR$ logic gates, as shown in Figure \ref{fig:computation-acceleration}.

\textbf{Number of bit-wise operations}. Compared with the previous approaches \cite{vo2022deep,fu2018towards}, in a binary convolution layer, the proposed method can better reduce the number of $\XNOR$s and parameters.
Firstly, as explained in the previous section, adding connections among centers leads to less computation for the K-mean approach \cite{vo2022deep} as in Eq. (\ref{eq:k-mean_compare}).
\vspace{-0.1cm}
\begin{equation}
     \label{eq:k-mean_compare}
     \sum_{j}d_{ij} + C_{in}\times M\times M < \sum_{j \not\in C}d_{ij} + R\times C_{in}\times M\times M,
\end{equation}
where $C$ includes all centers, $R$ is the number of centers.
Furthermore, adding connections among centers also creates a spanning tree, whose computational cost is always equal to or higher than using an MST.
Secondly, in \cite{fu2018towards}, the authors proposed using the shortest Hamiltonian path to arrange the computation order.
Because this path visits all vertices exactly once, the shortest Hamiltonian path is also a spanning tree.
Therefore, the number of $\XNOR$ bit-wise operations using an MST method is equal to or lower than the shortest Hamiltonian path.

\textbf{Exploration time}. The MST is faster and less complex for finding the computation order, compared to previous work \cite{fu2018towards,vo2022deep}.
Indeed, the MST is found by Prim's algorithm with the complexity of $O(V^{2})$ \cite{tutte2001graph}, where $V$ is the number of vertices.
In contrast, the K-mean algorithm requires $O(V^{2}G)$ \cite{hartigan1979algorithm} time complexity, where $G$ denotes the number of groups.
Since the optimal number of groups varies across convolutions, extra time is needed to determine this number for each case.
Meanwhile, the shortest Hamiltonian path has a much higher time complexity of $O(V^22^{V})$ \cite{bellman1962dynamic}, making it challenging to apply in practical settings.
To estimate these approaches for practical operation, we perform experiments for the first $6$ binary convolution layers of ResNet-20 \cite{he2016deep} with the weight data trained by \cite{xu2021recu}, while experiments on deeper layers with a larger number of output channels are infeasible even uncompleted for the K-mean and the shortest Hamiltonian path approach due to their longer running times.
According to Table \ref{tab:op-bit-time}, the MST approach gives the lowest bit-Ops and exploration time.

\begin{figure}[t]
\begin{center}
   \includegraphics[width=0.8\linewidth]{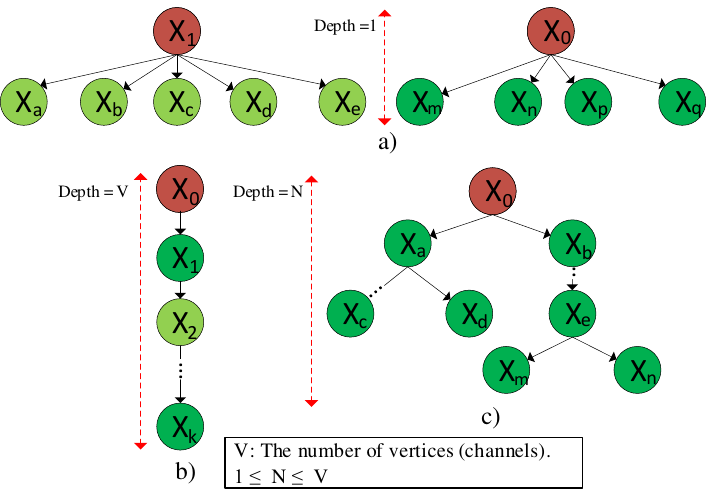}
\end{center}
\caption{a) The computation tree's depth using the K-mean cluster. b) The computation tree's depth using the shortest Hamiltonian path. c) The computation tree's depth using the MST.}
\label{fig:depth_tree}
\end{figure}

\begin{table}[]
\footnotesize
\begin{center}
\begin{tabular}{l|ll|ll|ll}
\hline
\multirow{2}{*}{\textbf{L}} & \multicolumn{2}{c|}{\textbf{Hamiltonian}} & \multicolumn{2}{c|}{\textbf{K-mean}} & \multicolumn{2}{c}{\textbf{MST}} \\
                            & Bit-Ops      & Time     & Bit-Ops   & Time   & Bit-Ops & Time \\ \hline
1                           & 1.08 M                & 31.3 s            & 1.33 M             & 0.090 s         & 1.07 M           & 0.021 s       \\
2                           & 1.09 M                & 32.9 s            & 1.30 M             & 0.075 s         & 1.06 M           & 0.021 s       \\
3                           & 1.07 M                & 32.6 s            & 1.30 M             & 0.085 s         & 1.05 M           & 0.023 s       \\
4                           & 1.10 M                & 32.7 s            & 1.33 M             & 0.076 s         & 1.08 M           & 0.021 s       \\
5                           & 1.11 M                & 32.5 s            & 1.32 M             & 0.115 s         & 1.09 M           & 0.020 s       \\
6                           & 1.09 M                & 32.5 s            & 1.33 M             & 0.086 s         & 1.08 M           & 0.021 s       \\ \hline
\end{tabular}
\end{center}
\caption{Bit-Ops and running time w.r.t. different approaches for reordering calculation on the first $6$ convolution layers of ResNet-20 with $16$ output channels.} 
\vspace{-0.3cm}
\label{tab:op-bit-time}
\end{table}

\textbf{Computation latency}. This is one of the factors affecting the resource overhead. 
This paper focuses on improving the architecture in \cite{vo2022deep}, a streaming acceleration for BNNs.
In doing so, the deeper the computation tree, the longer latency between the time output of the first and the last channel comes out.
Consequently, the number of added registers (Flip-flops) for synchronization tends to be equal or more for the deeper tree because all output channels must be available at the same clock for the next convolution operation.
The exact number of added registers depends on the number of adders executed in a clock period, which is affected by the required frequency, design complexity, and hardware platform.
Thus, to evaluate the comprehensive effects of the depth with practical implementations, the experiments related to this comparison are provided in Sec. \ref{sec:results}.
Figure \ref{fig:depth_tree} visually describes the depth of the MST, K-mean cluster, and the shortest Hamiltonian path approach.
Accordingly, the depths of the K-mean method and the shortest Hamiltonian path are always $1$ and $V$, respectively.
Meanwhile, the MST's depth can range from $1$ to $V$, depending on the graph structure.
Therefore, the proposed method is always better than the shortest Hamiltonian path in all aspects (number of operations, latency, running time).
Meanwhile, the K-mean method gives a better latency but much worse for the rest.
After exploring an MST for the post-training stage to limit the impact of the tree depth on latency and resources, we continue finding a new root that makes the depth of the new tree minimum.

\subsection{Learning Optimization} \label{sec:learning_op}
According to the observation mentioned earlier, to enhance the benefits of the MST approach, we propose a simple learning method that can further reduce the MST depth and total MST distance over a BNN model.
Specifically, to make the training converged, instead of directly optimizing the MST, we cluster weight sets ($\w_b^{li}$) to different groups and try to reduce the distance among weight sets in each group by optimizing the distances from all weight sets to fixed centers.
In doing so, given the $l^{th}$ binary convolution layer of a BNN model, we first randomly sample a binary center subset $\mathbb{C}_l \subset \{\pm1\}^{C_{in} \times M \times M}$ and $|\mathbb{C}_l| = N_l$.
Then, we explore the nearest center in $\mathbb{C}_l$ for every $\w_b^{li}$ and save it into a dictionary $\mathcal{C}$ using Eq. (\ref{eq:update_center}).
\vspace{-0.1cm}
\begin{equation}
     \label{eq:update_center}
     \mathcal{C}({\w}_b^{li}) = \underset{x}{\argmin}\left(\sum_{i=1}^{C_{in}\times M \times M} \lVert x- {\w}_b^{li}\rVert\right);
     x \in \mathbb{C}_l.
\end{equation}

On the other hand, the total distance from every weight sets to its center is added to the loss function and compels the training to minimize both the cross-entropy loss and the distance among weight sets in a group (distance loss).
To strike a balance between optimizing the original and distance loss, we adjust a hyper-parameter $\lambda$ to achieve an optimal ratio for different BNN models.
Besides, a $\gamma$ parameter serves as a hyper-parameter to control the priority of the original and additional loss during the training.
This parameter varies following a specific scheduler during the training process.
Particularly, the algorithm prioritizes reducing the MST distance and depth during the initial training phase and gradually shifting its focus toward accuracy at the end of the process.
The new loss function is shown in Eq. (\ref{eq:loss}), where $\mathcal{L}_0(\inputs, {\w})$ is the original loss function.
Forward processes are summarized in Algorithm \ref{alg:full_alg}.
\begin{equation}
     \label{eq:loss}
     \mathcal{L}(\inputs,{\w}) = \mathcal{L}_0(\inputs, {\w}) + \lambda\gamma\sum_i \lVert\mathcal{C}({\w}_b^{i}) - {\w}_b^{i} \rVert^2,
\end{equation}

\begin{algorithm}
\footnotesize
 \caption{Forward propagation for the training process.} \label{alg:full_alg}
 \begin{algorithmic}[1]
 \renewcommand{\algorithmicrequire}{\textbf{Require:}}
 \REQUIRE  Input data, pre-train full-precision weight ($\w$), threshold ($\theta$), learning rate ($\eta$), MST scaling factor ($\lambda$).
  \STATE Load pre-train weight ($\w$)
  \FOR {epoch = 1 $\rightarrow$ $E$} 
  \STATE $\mathcal{L}_{mst}$ = 0
  \FOR {l = 1 to L} 
      \STATE ${\w}_b^l$ = $\Sign({\w}_l$) \cite{xu2021recu}
      \STATE $\text{a}_b^{l-1}$ = $\Sign(\text{a}_{l-1}$)
      \FOR {i = $1$ to $C_{out}$}
          \STATE Update centers for ${\w}_b^{l}$:\\
          \STATE $\mathcal{C}({\w}_b^{li})$ = $\underset{x}{\argmin}(\sum_{i=1}^{C_{in}\times M\times M} \lVert x- {\w}_b^{li}\rVert$)  $; x \in \mathbb{C}_l$ 
          \STATE $\mathcal{L}_{MST} = \mathcal{L}_{MST} + \sum \lVert\mathcal{C}({\w}_b^{li}) - {\w}_b^{li} \rVert^2$
      \ENDFOR
      \STATE $\text{a}_l$ = $({\w}_b^l \circledast \text{a}_b^{l-1}) \odot \alpha_l$
  \ENDFOR
  \STATE Compute total loss: $\mathcal{L} = \mathcal{L}_0 + \lambda*\gamma*\mathcal{L}_{MST}$
  \ENDFOR
 \end{algorithmic} 
 \end{algorithm}
 
It is noticeable that we use pre-trained BNN models for the learning process to avoid diverging during distance optimization time.
In this paper, results from ReCu \cite{xu2021recu} are used and considered the baseline for comparison.

\subsection{Streaming Acceleration} \label{sec:hw_architecture}
\begin{figure}[t]
\begin{center}
   \includegraphics[width=1\linewidth]{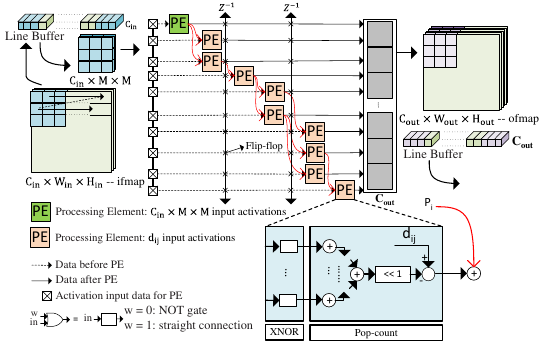}
\end{center}
\caption{High-speed hardware accelerator for a binary convolution layer, where the $\XNOR$s are simplified with $\NOTs$ or simple straight connection, while computation on each PE corresponding to each output channel is executed gradually with the MST.}
\vspace{-0.3cm}
\label{fig:hw-acceleration}
\end{figure}

Inspired by the streaming architecture in \cite{vo2022deep}, this paper proposes an entire BNN high-speed architecture for BNN models applying our compression method.
Specifically, the hardware architecture of a binary convolution layer is shown in Figure \ref{fig:hw-acceleration}, in which the process is divided into two steps: 1) loading the input feature map to a line buffer and 2) executing $\XNOR+\Popcount$ operation.

\textbf{Loading input.}
Given a binary convolution layer with $M = 3$, padding size $= 1$, firstly, $C_{in}$ input feature maps are gradually loaded into $C_{in}$ line buffers with the size $2\times W_{in} + 3$.
When $W_{in} + 2$ registers in each line buffer are filled, $C_{in}\times 3\times 3$ input pixels are selected from the line buffers for the convolution ($\XNOR + \Popcount$) operation.
In standard calculation approach, the number of $\XNOR$ operations used for a binary convolution is $C_{out}\times C_{in}\times M\times M$.
These operations can be separated into $C_{out}$ sub-operations corresponding to $C_{out}$ output channels; each sub-operation includes $C_{in}\times M \times M \XNOR$ bit-operations and requires $C_{in}\times M\times M$ corresponding input pixels and weight values.
In this proposed architecture, the difference is that the number of inputs for a particular sub-operation is lower than $C_{in}\times M\times M$ and equal to the distance between the corresponding vertex and parent vertex as in the constructed graph, except the sub-operation for the root vertex requiring full calculation with $C_{in}\times M\times M \XNOR$s.

\textbf{Perform convolution.} This step executes the convolution operation with the order of calculation following the MST.
As shown in Figure \ref{fig:hw-acceleration}, the output of a specific PE (red arrow) is reused for the next PE to reduce computational cost.
The PE for the output channel $i^{th}$ needs $d_{ij}$ weight values and executes $d_{ij}$ $\XNOR$ operations with the corresponding $\Popcount$ operation, where $j$ is the parent vertex of the vertex $i^{th}$.
Besides, to eliminate memory for convolution weight storage, $\XNOR$ operations are simplified to NOT gate or straight connection.
In terms of timing violations, we pipeline the $\Popcount$ operation by adding register lines among the adder tree to reduce computation complexity in a clock cycle.
The exact number of register lines depends on the MST depth, frequency and hardware platform.
Thus, depending on practical implementations, we adjust this number to meet the timing requirements.

\section{Experiments} \label{sec:results}
In this section, we conduct experiments including software and hardware implementation to show the effectiveness of the proposed method for BNN compression.
Training experiments are executed with the popular datasets CIFAR-10 and ImageNet via Pytorch, while hardware implementation is performed on the Xilinx FPGA platforms.

\subsection{Learning and hardware implementation}
\textbf{Training implementation}: 
Because our proposed method is independent of other BNN accuracy optimization methods, in this paper, we select training results from the ReCU approach in \cite{xu2021recu} as a case study for further compressing.
In doing so, VGG-small, ResNet-18/20 models are used to evaluate on CIFAR-10 dataset, while ResNet-18/34 are used for ImageNet.
Regarding training in detail, we keep the configuration as in \cite{xu2021recu}, except for the learning rate of 0.1.
In addition, the $\lambda$ value is adjusted depending on the training models, datasets, and focused objectives.
Meanwhile, the $\gamma$ parameter is scheduled following the learning rate.
In this paper, when training with CIFAR-10, we select 1e-6 for ResNet-18, 5e-6 for ResNet-20, and 4e-6 for VGG-small.
Meanwhile, for ImageNet, we select 1e-6 for ResNet-18 and 1e3 for ResNet-34.

\textbf{Hardware implementation}: 
The inference hardware architecture is implemented on a Xilinx FPGA platform called xczu3eg-sbva484 part.
Before programming to FPGA, the design is run with the corresponding C/C++ model and simulated via Synopsys VCS.
Finally, Vivado 2018.3 is used for  the design synthesis and implementation.
\subsection{Ablation Studies}
In this section, experimental results and discussion related to hyper-parameter $\lambda$, and $\gamma$ are described.

\begin{figure}[t]
\centering
        \begin{tikzpicture}
        \begin{groupplot}[group style={group size=2 by 1,horizontal sep = 1cm, vertical sep = 1.2cm},height=4cm,width=4.8cm,]
        \nextgroupplot[
            xlabel={a) MST distances of each layer (L).},
            xlabel style={font=\footnotesize},
            xmin=0, xmax=600,
            ytick={50000,150000,250000,350000},
            every tick label/.append style={font=\footnotesize},
            legend pos=south east,
            legend image post style={scale=0.15},
            every axis legend/.append style={font=\footnotesize,at={(0.37,0.97)}, anchor=north west,legend columns = 3},
            ymajorgrids=true,     xmajorgrids=true,
            grid style=dashed,
            ylabel={MST distance},
            ylabel style={font=\footnotesize, yshift=-.5cm},
            ymax=400000, ymin= 0,
         ]
                \addplot [color=blue,  line width=0.8pt]
                table {TIKZ/data/L1.dat}; %
                \addplot [color=green, line width=0.8pt]
                table {TIKZ/data/L2.dat}; %
                \addplot [color=black,  line width=0.8pt]
                table {TIKZ/data/L3.dat};
            \addplot [color=red,  line width=0.8pt]
                table {TIKZ/data/L4.dat};
            \addplot [color=brown,  line width=0.8pt]
                table {TIKZ/data/L5.dat};
                \legend{L1, L2, L3, L4, L5}

        \nextgroupplot[
            xlabel={b) Accuracy curves.},
            xmin=0, xmax=600,
            ymajorgrids=true,     xmajorgrids=true,
            legend pos=south east,
            legend image post style={scale=0.2},
            every tick label/.append style={font=\footnotesize},
            every axis legend/.append style={font=\footnotesize},
            grid style=dashed,
            ylabel={Accuracy},
            ylabel style={yshift=-.5cm},
            every axis label/.append style={font=\footnotesize},
            ymax=100, ymin= 0,
        ]
            \addplot [color=blue,line width=0.8pt]
            table {TIKZ/data-loss/train_error.tex}; %
            \addplot [color=green,line width=0.8pt]
            table {TIKZ/data-loss/val_error.tex}; %
            \legend{Training, Validation}
            \end{groupplot}
        \end{tikzpicture}

\caption{MST distance and top-1 $\%$ accuracy during the training process, where the Binary VGG-small model and CIFAR-10 dataset is used for the training experiment with 600 epochs.}
\vspace{-0.4cm}
\label{fig:training-loss}
\end{figure}
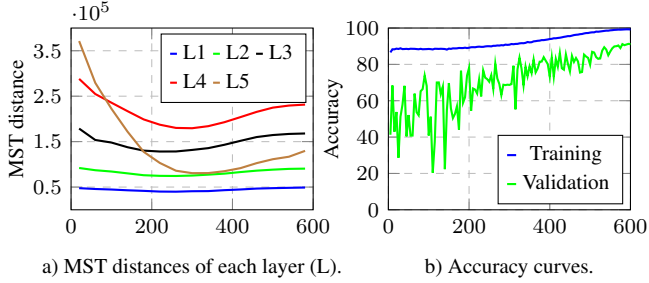

\textbf{Effect of hyper-parameter $\gamma$}: As explained in Sec. \ref{sec:sub-MST_inference}, an additional loss related to MST distance is added for the learning.
The effect of the loss depends on the $\gamma$, while the $\gamma$ changes following the epoch.
To facilitate the evaluation, Figure \ref{fig:training-loss} provides MST distance curves for all binary convolution layers and training-validation accuracy curves in the entire learning process.
In the first half-time, the $\gamma$ is still significant, and the training focuses on MST distance optimization, while the accuracy optimization is temporarily less affected.
Consequently, the MST distance of all binary convolutions is significantly down, especially for the deeper layers with more channels, while the training accuracy suddenly decreases and the validation accuracy goes down and strongly fluctuates at this time.
In the second half-time, the training focuses on accuracy optimization with the lower $\gamma$ value.
Thus, the training accuracy tends to increase, while the validation accuracy rises, too, with a lower fluctuation.
Although the MST distance also escalates again on all layers, these values are lower than at the initial time, especially on deeper layers.
This is the reason why we can save computational costs by using this approach.

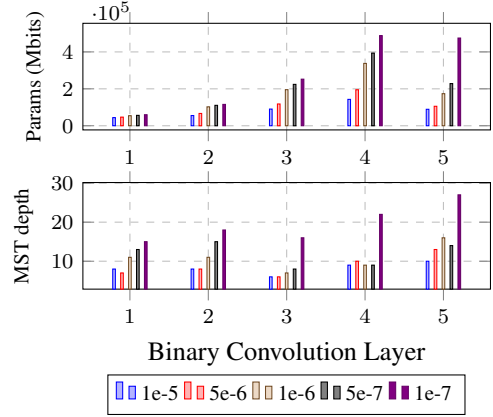
\begin{figure}[t]
\begin{center}
        \pgfplotsset{compat=newest}
        \begin{tikzpicture}
        \begin{groupplot}[group style={group size=1 by 2,horizontal sep = 1.2cm, vertical sep = 0.7cm},height=3cm,width=7cm, ]
        \nextgroupplot[
        	x tick label style={/pgf/number format/1000 sep=},
        	ylabel=Params (Mbits),
        	enlargelimits=0.15, grid=both,
        	legend style={at={(0.5,-0.5)},
        		anchor=north,legend columns=-1},
                every tick label/.append style={font=\footnotesize},
                every axis legend/.append style={font=\footnotesize},
                ylabel style={font=\footnotesize},
        	ybar,
                grid style=dashed,
        	bar width=1pt,
        ]
        \addplot
        	coordinates {(1,44071) (2,55358)
        		 (3,90383) (4,142927) (5,89215)};
        \addplot
        	coordinates {(1,46856) (2,66685)
        		 (3,117990) (4,195030) (5,105542)};
        \addplot
        	coordinates {(1,54576) (2,102437)
        		 (3,195534) (4,337708) (5,173838)};
        \addplot
        	coordinates {(1,56353) (2,110681)
        		 (3,224240) (4,393732) (5,228356)};
        \addplot
        	coordinates {(1,59719) (2,115705)
        		 (3,252384) (4,488311) (5,475170)};
        
        \nextgroupplot[
        	x tick label style={/pgf/number format/1000 sep},
        	ylabel=MST depth,
        	enlargelimits=0.15, grid=both,
        	legend style={at={(0.5,-0.7)},anchor=north,legend columns=-1},
                every tick label/.append style={font=\footnotesize},
                every axis legend/.append style={font=\footnotesize},
        	ybar,
                grid style=dashed,
        	bar width=1pt,
                xlabel= Binary Convolution Layer,
                legend image post style={scale=0.8},
                every axis legend/.append style={font=\footnotesize,at={(0.06,-0.8)}, anchor=north west,legend columns = 5},
                ylabel style={font=\footnotesize},
        ]
        \addplot
        	coordinates {(1,8) (2,8)
        		 (3,6) (4,9) (5,10)};
        \addplot
        	coordinates {(1,7) (2,8)
        		 (3,6) (4,10) (5,13)};

        \addplot
        	coordinates {(1,11) (2,11)
        		 (3,7) (4,9) (5,16)};
        \addplot
        	coordinates {(1,13) (2,15)
        		 (3,8) (4,9) (5,14)};
        \addplot
        	coordinates {(1,15) (2,18)
        		 (3,16) (4,22) (5,27)};

        \legend{1e-5,5e-6,1e-6, 5e-7, 1e-7}
        \end{groupplot}
        \end{tikzpicture}
    
\end{center}
\caption{Number of parameters and MST depth on each convolution layer \textit{w.r.t.} different $\lambda$ values.}
\label{fig:lamda}
\end{figure}

\begin{table}[]
\begin{center}
\footnotesize
\begin{tabular}{lcccc}
\hline

$\lambda$&\multicolumn{1}{l}{MST-depth}      & \multicolumn{1}{l}{\begin{tabular}[c]{@{}l@{}}\#Params\\ (Mbit)\end{tabular}} & \multicolumn{1}{l}{\begin{tabular}[c]{@{}l@{}}\#Bit-Ops\\ (GOps)\end{tabular}} & \multicolumn{1}{c}{\begin{tabular}[c]{@{}c@{}}Top-1 Acc.\\ mean $\pm$ std (\%)\end{tabular}} \\ \hline
1e-7      & $97.3$         & $1.391$             & $0.217$              & $92.17\pm 0.07$              \\ \hline
5e-7      & $58.3$         & $0.998$             & $0.184$              & $92.09\pm 0.06$              \\ \hline
1e-6      & $54.0$         & $0.864$             & $0.165$              & $91.99\pm 0.07$              \\ \hline
5e-6      & $44.0$         & $0.532$             & $0.115$              & $91.14\pm 0.08$              \\ \hline
1e-5      & $42.3$         & $0.422$             & $0.098$              & $90.17\pm 0.14$              \\ \hline

\end{tabular}
\end{center}
\caption{MST depth, number of parameters, bit-Ops, and accuracy \textit{w.r.t.} different values of $\lambda$ on CIFAR-10 VGG-small model.}
\vspace{-0.3cm}
\label{tab:comparison_lamda}
\end{table}

\textbf{Effect of hyper-parameter $\lambda$}: The hyper-parameter $\lambda$ helps scale down the effect of the total MST distance to the new loss function.
Each model has a different number of layers and channels that directly affect the total MST distance.
Therefore, depending on the model, we adjust this hyper-parameter to balance the loss for accuracy and MST distance.
In this section, the VGG-small model and CIFAR-10 dataset are used to estimate how the $\lambda$ impacts the number of parameters, bit-operations, MST-depth, and accuracy with the range of $\lambda$ from 1e-7 to 1e-5.
Table \ref{tab:comparison_lamda} and Figure \ref{fig:lamda} provide the number of parameters, bit-operations, MST-depth, accuracy, and the number parameters for each binary convolution layer \textit{w.r.t} different $\lambda$ values.
Accordingly, the number of parameters, bit-operations, and accuracy tend to be lower with higher $\lambda$ values.
However, the number of parameters/bit-ops is considerably reduced at $3.3\times$, while the accuracy degradation is acceptable with $2$\% when increasing the $\lambda$ from 1e-7 to 1e-5.
Besides, the MST depth also decreases when the $\lambda$ value increases.
Especially, when increasing $\lambda$ from 1e-7 to 5e-7, the depth reduce $1.67\times$.

\subsection{Comparison with SOTA methods}
To demonstrate the effectiveness of the proposed method, we perform a series of training experiments with various NN models on CIFAR-10, and ImageNet, and then compare them with state-of-the-art methods.
Specifically, for each NN model, we provide two experimental results corresponding to two cases 1) apply both the learning optimization in Sec. \ref{sec:learning_op} and the MST reordering to the final training result (\textbf{Ours 1}); 2) directly apply the MST reordering to pre-trained BNN models (\textbf{Ours 2}).
In this paper, the pre-trained BNN models we use for fine-tuning and exploring the MST are provided by Xu \etal \cite{xu2021recu}.
In terms of acceleration, a neural network model is fully trained on CIFAR-10 and implemented on an FPGA hardware platform to compare with recent hardware architectures.

\begin{table}[]
\begin{center}
\begin{threeparttable}[t]
\footnotesize
\begin{tabular}{l|lccc}
\hline
N-work&\multicolumn{1}{l|}{Method}                               & \multicolumn{1}{l|}{\begin{tabular}[c]{@{}l@{}}\#Params\\ (Mbit)\end{tabular}} & \multicolumn{1}{l|}{\begin{tabular}[c]{@{}l@{}}\#Bit-Ops\\ (GOps)\end{tabular}} & \multicolumn{1}{c}{\begin{tabular}[c]{@{}c@{}}Top-1 Acc.\\(\%)\end{tabular}} \\ \hline
\multirow{7}{*}{\begin{tabular}[c]{@{}l@{}}VGG\\ small\end{tabular}}   &\multicolumn{1}{l|}{RAD\cite{ding2019regularizing}}                                  & \multicolumn{1}{c|}{4.571}                                                     & \multicolumn{1}{c|}{0.603}                                 & 90.0                                 \\
    &\multicolumn{1}{l|}{IR-Net\cite{qin2020forward}}                               & \multicolumn{1}{c|}{4.571}                                                     & \multicolumn{1}{c|}{0.603}                                 & 90.4                                 \\
    &\multicolumn{1}{l|}{RBNN\cite{lin2020rotated}}                                 & \multicolumn{1}{c|}{4.571}                                                     & \multicolumn{1}{c|}{0.603}                                 & 91.3                                 \\
    &\multicolumn{1}{l|}{Adabin\cite{tu2022adabin}}                               & \multicolumn{1}{c|}{4.571}                                                     & \multicolumn{1}{c|}{0.603}                                 & 92.3                                 \\
    &\multicolumn{1}{l|}{ReCU\cite{xu2021recu}}                                 & \multicolumn{1}{c|}{4.571}                                                     & \multicolumn{1}{c|}{0.603}                                 & 92.4                                 \\
    &\multicolumn{1}{l|}{SNN\cite{wang2021snn}}                                  & \multicolumn{1}{c|}{3.047}                                                     & \multicolumn{1}{c|}{0.194}                                 & 91.0                                 \\
    &\multicolumn{1}{l|}{\textbf{Ours 1$|$2}}                        & \multicolumn{1}{c|}{\text{0.556$|$1.611}}                                            & \multicolumn{1}{c|}{\text{0.122$|$0.232}}                        & 91.5$|$93.3\tnote{*} \\
   \hline
\multirow{7}{*}{\begin{tabular}[c]{@{}c@{}}ResNet\\ 18\end{tabular}}     &\multicolumn{1}{l|}{RAD\cite{ding2019regularizing}}                                  & \multicolumn{1}{c|}{10.99}                                                     & \multicolumn{1}{c|}{0.547}                                 & 90.5                                 \\
    &\multicolumn{1}{l|}{IR-Net\cite{qin2020forward}}                               & \multicolumn{1}{c|}{10.99}                                                     & \multicolumn{1}{c|}{0.547}                                 & 91.5                                 \\
    &\multicolumn{1}{l|}{RBNN\cite{lin2020rotated}}                                 & \multicolumn{1}{c|}{10.99}                                                     & \multicolumn{1}{c|}{0.547}                                 & 92.2                                 \\
    &\multicolumn{1}{l|}{ReCU\cite{xu2021recu}}                                 & \multicolumn{1}{c|}{10.99}                                                     & \multicolumn{1}{c|}{0.547}                                 & 92.8                                 \\
    &\multicolumn{1}{l|}{Adabin\cite{tu2022adabin}}                               & \multicolumn{1}{c|}{10.99}                                                     & \multicolumn{1}{c|}{0.547}                                 & 93.1                                 \\
    &\multicolumn{1}{l|}{SNN\cite{xu2021recu}}                                  & \multicolumn{1}{c|}{7.324}                                                     & \multicolumn{1}{c|}{0.289}                                 & 91.0                                 \\
    &\multicolumn{1}{l|}{\textbf{Ours 1$|$2}} & \multicolumn{1}{c|}{{\text{0.814$|$4.293}}}                     & \multicolumn{1}{c|}{{ \text{0.104$|$0.216}}} & { 91.6$|$93.2\tnote{*}} \\
   \hline
\multirow{7}{*}{\begin{tabular}[c]{@{}c@{}}ResNet\\ 20\end{tabular}}     &\multicolumn{1}{l|}{DSQ\cite{gong2019differentiable}}                                  & \multicolumn{1}{c|}{0.267}                                                     & \multicolumn{1}{c|}{0.040}                                 & 84.1                                 \\
    &\multicolumn{1}{l|}{IR-Net\cite{qin2020forward}}                               & \multicolumn{1}{c|}{0.267}                                                     & \multicolumn{1}{c|}{0.040}                                 & 86.5                                 \\
    &\multicolumn{1}{l|}{RBNN\cite{lin2020rotated}}                                 & \multicolumn{1}{c|}{0.267}                                                     & \multicolumn{1}{c|}{0.040}                                 & 86.5                                 \\
    &\multicolumn{1}{l|}{ReCU\cite{xu2021recu}}                                 & \multicolumn{1}{c|}{0.267}                                                     & \multicolumn{1}{c|}{0.040}                                 & 87.4                                 \\
    &\multicolumn{1}{l|}{Adabin\cite{tu2022adabin}}                               & \multicolumn{1}{c|}{0.267}                                                     & \multicolumn{1}{c|}{0.040}                                 & 88.2                                 \\
    &\multicolumn{1}{l|}{{SNN\cite{wang2021snn}}}           & \multicolumn{1}{c|}{{0.178}}                              & \multicolumn{1}{c|}{{0.040}}          & {85.1}          \\
    &\multicolumn{1}{l|}{{\textbf{Ours 1$|$2}}} & \multicolumn{1}{c|}{{{0.096$|$0.116}}}                     & \multicolumn{1}{c|}{{ \text{0.015$|$0.017}}} & {86.5$|$88.0\tnote{*}}\\
 \hline
\end{tabular}
   \begin{tablenotes}
     \item[*] Accuracy after fine-tuning is at https://github.com/z-hXu/ReCU \cite{xu2021recu}.
   \end{tablenotes}
\end{threeparttable}%
\end{center}
\caption{ 
Comparison with the state-of-the-art methods on CIFAR-10. The bit-width is 1 for both activation and weight.}
\label{tab:comparison_sota_cifar}
\end{table}

\textbf{Software training comparison}: For CIFAR-10, we conduct experiments on three neural BNN models: RestNet-18/20 and VGG-small.
we compare our results with RAD \cite{ding2019regularizing}, IR-Net \cite{qin2020forward}, Adabin \cite{tu2022adabin}, ReCU \cite{xu2021recu}, DSQ \cite{gong2019differentiable}, and SNN \cite{wang2021snn}.
Specifically, as shown in Table \ref{tab:comparison_sota_cifar}, when comparing \textbf{Ours 1} with the highest-accuracy methods on all models, our proposed method reduces $8.2\times$ parameters and $4.9\times$ bit-Ops with a $0.97$\% accuracy drop on VGG-small.
Meanwhile, when directly exploring the MST on the pre-trained models, \textbf{Ours 2} gets $2.8\times$ parameters and $2.6\times$ bit-Ops reduction without accuracy degradation. 
On ResNet-18, \textbf{Ours 1} gives $13.50\times$ parameters, $5.26\times$ bit-Ops reduction with a $1.29$\% accuracy drop, while \textbf{Ours 2} reduces $2.6\times$ parameters and $2.5\times$ bit-Ops without compromising accuracy. 
Additionally, we have $2.78\times$ parameters, $2.67\times$ bit-Ops reduction, and a $1$\% accuracy drop with \textbf{Ours 1}; $2.3\times$ parameters, $2.35\times$ bit-Ops reduction and $0\%$ accuracy drop on ResNet-20. 

Compared \textbf{Ours 1} with the compression method SNN \cite{wang2021snn} on all models, the proposed method obtains a better compression ratio with a maximum reduction of $9.0\times$ parameters, $2.78\times$ bit-Ops on ResNet-18 while yielding higher accuracy and up to maximum $1.4$\% on ResNet-20.
Meanwhile, with \textbf{Our 2}, we reduce maximum $1.89\times$ parameters on VGG-small, $2.35\times$ bit-Ops on ResNet-20, while achieving higher accuracy up to maximum $2.9\%$ on ResNet-20.

For ImageNet, experiments are implemented on ResNet-18/34 and comparison is performed with BNN+ \cite{hubara2016binarized}, Bi-Real \cite{liu2018bi}, XNOR++ \cite{bulat2019xnor}, IR-Net \cite{qin2020forward}, Adabin \cite{tu2022adabin}, ReCU \cite{xu2021recu}, SNN \cite{wang2021snn}.
According to Table \ref{tab:comparison_sota_imagenet}, \textbf{Ours 1} reduces $3.2\times$ parameters, $2.6\times$ bit-Ops on ResNet-18; and $2.23\times$ parameters, $2.27\times$ bit-Ops on ResNet-34, compared to the baseline \cite{xu2021recu}, while the accuracy degradation is $4$\% and $2.2$\% on ResNet18 and ResNet-34, respectively.
\textbf{Ours 2} achieves $2.27\times$ parameters, $2.34\times$ bit-Ops reduction on ResNet-18; and $2.2\times$ parameters, $2.26\times$ bit-Ops reduction without accuracy drop. 
Compared to SNN with the highest compression ratio recently, our proposed method with \textbf{Ours 1} saves $2.13\times$ parameters and $1.3\times$ bit-Ops on ResNet-18, $1.49\times$ parameters and $1.09\times$ bit-Ops on ResNet-34, while the trained models give higher accuracy of $0.7$\% and $1.5$\% on ResNet-18, ResNet-34, respectively.
Meanwhile, \textbf{Ours 2} without accuracy drop saves up to $1.51/1.47\times$ parameters and $1.23/1.08\times$ bit-Ops on ResNet-18/34, respectively.

\begin{table}[]
\footnotesize
\begin{center}
\begin{threeparttable}%
\begin{tabular}{l|lccc}
\hline
N-work&\multicolumn{1}{l|}{Method}                               & \multicolumn{1}{c|}{\begin{tabular}[c]{@{}l@{}}\#Params\\ (Mbit)\end{tabular}} & \multicolumn{1}{c|}{\begin{tabular}[c]{@{}l@{}}\#Bit-Ops\\ (GOps)\end{tabular}} & \multicolumn{1}{c}{\begin{tabular}[c]{@{}c@{}}Top-1 Acc.\\ (\%)\end{tabular}} \\ \hline
\multirow{8}{*}{\begin{tabular}[c]{@{}c@{}}ResNet\\ 18\end{tabular}}&\multicolumn{1}{l|}{BNN+\cite{hubara2016binarized}}                                 & \multicolumn{1}{c|}{10.99}                                                     & \multicolumn{1}{c|}{1.677}                                                      & 53.0                                                                          \\
&\multicolumn{1}{l|}{Bi-Real\cite{liu2018bi}}                              & \multicolumn{1}{c|}{10.99}                                                     & \multicolumn{1}{c|}{1.677}                                                      & 56.4                                                                          \\
&\multicolumn{1}{l|}{XNOR++\cite{bulat2019xnor}}                               & \multicolumn{1}{c|}{10.99}                                                     & \multicolumn{1}{c|}{1.677}                                                      & 57.1                                                                          \\
&\multicolumn{1}{l|}{IR-Net\cite{qin2020forward}}                               & \multicolumn{1}{c|}{10.99}                                                     & \multicolumn{1}{c|}{1.677}                                                      & 58.1                                                                          \\
&\multicolumn{1}{l|}{Adabin\cite{tu2022adabin}}                               & \multicolumn{1}{c|}{10.99}                                                     & \multicolumn{1}{c|}{1.677}                                                      & 63.1                                                                          \\
&\multicolumn{1}{l|}{ReCU\cite{xu2021recu}}                                 & \multicolumn{1}{c|}{10.99}                                                     & \multicolumn{1}{c|}{1.677}                                                      & 61.0                                                                          \\
&\multicolumn{1}{l|}{SNN\cite{wang2021snn}}                                  & \multicolumn{1}{c|}{7.32}                                                      & \multicolumn{1}{c|}{0.883}                                                      & 56.3                                                                          \\
&\multicolumn{1}{l|}{{ \textbf{Ours 1$|$2}}} & \multicolumn{1}{c|}{{ \text{3.43$|$4.84}}}                     & \multicolumn{1}{c|}{{ \text{0.636$|$0.716}}}                      & { 57.0$|$61.2\tnote{*}}                                        \\
 \hline
\multirow{6}{*}{\begin{tabular}[c]{@{}c@{}}ResNet\\ 34\end{tabular}}&\multicolumn{1}{l|}{Bi-Real\cite{liu2018bi}}                              & \multicolumn{1}{c|}{21.09}                                                     & \multicolumn{1}{c|}{3.526}                                                      & 62.2                                                                          \\
&\multicolumn{1}{l|}{IR-Net\cite{qin2020forward}}                               & \multicolumn{1}{c|}{21.09}                                                     & \multicolumn{1}{c|}{3.526}                                                      & 62.9                                                                          \\
&\multicolumn{1}{l|}{Adabin\cite{tu2022adabin}}                               & \multicolumn{1}{c|}{21.09}                                                     & \multicolumn{1}{c|}{3.526}                                                      & 66.4                                                                          \\
&\multicolumn{1}{l|}{ReCU\cite{xu2021recu}}                                 & \multicolumn{1}{c|}{21.09}                                                     & \multicolumn{1}{c|}{3.526}                                                      & 65.1                                                                          \\
&\multicolumn{1}{l|}{SNN\cite{wang2021snn}}                                  & \multicolumn{1}{c|}{14.06}                                                     & \multicolumn{1}{c|}{1.696}                                                      & 61.4                                                                          \\
&\multicolumn{1}{l|}{{ \textbf{Ours 1$|$2}}} & \multicolumn{1}{c|}{{\text{9.44$|$9.51}}}                         & \multicolumn{1}{c|}{{ \text{1.550$|$1.558}}}                          & { 62.9$|$65.4\tnote{*}}                                                                           \\ \hline
\end{tabular}
   \begin{tablenotes}
     \item[*] Accuracy after fine-tuning is at https://github.com/z-hXu/ReCU \cite{xu2021recu}.
   \end{tablenotes}
\end{threeparttable}%
\end{center}
\caption{
Comparison with the state-of-the-art methods on ImageNet. The bit-width is 1 for both activation and weight. }
\label{tab:comparison_sota_imagenet}
\end{table}

\textbf{Hardware implementation comparison}: Table \ref{tab:hardware_comparison} provides the hardware performance comparison with previous works using CIFAR-10 for the training.
we compare the proposed architecture with FINN \cite{umuroglu2017finn}, FINN-R \cite{blott2018finn} from Xilinx, ReBNet \cite{ghasemzadeh2018rebnet} and the design in \cite{vo2022deep}.
Accordingly, the proposed accelerator takes the leading place in both speed and area efficiency (FPS/LUTs).
More specifically, it performs $1.6\times$ faster than the fastest previous work (FINN), while obtaining $3.7\times$ area efficiency with a little higher accuracy.
In addition, compared to \cite{vo2022deep}, our method gains over $1.81\times$ area efficiency.
Besides, to compare our method with the K-mean approach, we implement another design from the same training result with the K-mean method.
According to Table \ref{tab:hardware_comparison}, our method performs better $1.25\times$ in terms of resources and area efficiency.

\begin{table}[]
\footnotesize
\begin{center}
\begin{tabular}{l|ccccc}
\hline
Design & \begin{tabular}[c]{@{}c@{}}Freq\\ (MHz)\end{tabular} & LUTs    & \begin{tabular}[c]{@{}c@{}}Acc.\\ (\%)\end{tabular}  & \begin{tabular}[c]{@{}c@{}}FPS\\ (K)\end{tabular} & \begin{tabular}[c]{@{}c@{}}FPS/\\ LUTs\end{tabular} \\ \hline
FINN\cite{umuroglu2017finn}   & 200                                                  & 46,253  & 80.1  & 22                                              & 0.47     \\ 
FINN\cite{umuroglu2017finn}   & 125                                                  & 365,963 & 80.1  & 128                                               & 0.35     \\ 
FINN-R\cite{blott2018finn} & 237                                                  & 332,637 & 80.1  & 105                                             & 0.31     \\ 
FINN-R\cite{blott2018finn} & 300                                                  & 41,733  & 80.1  & 20                                                & 0.48     \\ 
FINN\cite{blott2018finn}   & 300                                                  & 25,431  & 80.1  & 1.9                                               & 0.07     \\ 
ReBNet\cite{ghasemzadeh2018rebnet} & 200                                                  & 53,200  & 80.6 & 6                                                 & 0.11     \\ 
\cite{vo2022deep}   & 210                                                  & 290,012 & 80.2  & 205                                               & 0.70     \\ 
Ours (K-mean)   & 210                                                  & 201,434 & 80.5  & 205                                               & 1.01     \\ 
\textbf{Ours (MST)}   & \textbf{210}                                                  & \textbf{161,294} & \textbf{80.5}  & \textbf{205}                                               & \textbf{1.27}     \\ \hline
\end{tabular}
\end{center}
\caption{Hardware performance comparison with the state-of-the-art architectures on CIFAR-10.}\
\vspace{-0.6cm}
\label{tab:hardware_comparison}
\end{table}
\section{Conclusion}
In conclusion, this paper introduces a comprehensive method for compressing BNNs, which covers the entire process from learning to hardware implementation. By utilizing the minimum spanning tree (MST), we effectively reduce the computation cost of binary convolution calculation. To optimize the MST during the training stage, a learning algorithm is proposed. Moreover, we present a hardware implementation for the inference task. Our experimental results demonstrate that the proposed method significantly reduces computation costs while maintaining high accuracy, making it promising for practical applications.
\section{Acknowledgements}
This work was supported by the NRF grant funded by the Korea government(MSIT) (No. RS-2023-00207816), and part by IITP Grant funded by MSIT (Artificial Intelligence Innovation Hub) under Grant 2021-0-02068, (No. RS-2022-00155911, AI Convergence Innovation Human Resources Development (Kyung Hee University)), IITP No.2019-0-01287, Evolvable Deep Learning Model Generation Platform for Edge Computing and ITRC support program (IITP-2023-RS-2023-00258649).

{\small
\bibliographystyle{ieee_fullname}
\bibliography{egbib}
}
\textbf{\large{Appendix}}
\newline

\textbf{Proof of Eq. (\ref{eq:conv_final})}

According to the convolution definition, the output of the channel $i$ as the following
\begin{equation} \tag{8}
\label{eq:conv}
Y_i = \left(\sum_{j=1}^{C_{in}MM}\mathcal{A}^b_{ij}*\mathcal{W}^b_{ij}\right) \odot \alpha
\end{equation}

There are $C_{in} \times M\times M$ multiplications, and these multiplications' output is -1 or 1.
Assuming that there are $A$ multiplications with the output $-1$ and $B$ multiplication with output $1$, we have $A + B = C_{in} \times M \times M$.
Thus, $Y_i$ can be derived as

\begin{equation} 
\label{eq:conv_1}
Y_i = A - B = 2A - C_{in} \times M \times M.
\end{equation}

In addition, because $\mathcal{A}^b_{ij}$ and $\mathcal{W}^b_{ij}$ are binarized, A can be calculated as

\begin{equation} 
\label{eq:conv_2}
A = \sum_{j=1}^{C_{in}MM}\XNOR (\mathcal{A}^b_{ij},\mathcal{W}^b_{ij}).
\end{equation}

Finally, we have Eq. (\ref{eq:conv_final}) by replacing A with Eq. (\ref{eq:conv_2}) as
\begin{equation} \tag{2}
\label{eq:conv_final}
Y_i = (2\sum_{j=1}^{C_{in}MM}\XNOR(\mathcal{A}^b_{ij},\mathcal{W}^b_{ij}) - C_{in}\times M\times M) \odot \alpha.
\end{equation}
\textbf{Proof of Eq. (\ref{eq:output_final})}

Assuming that $\mathcal{S}$ includes weight values of the channel $j$, which are similar to the weights of the channel $i$ (compared one-one respectively).
$\mathcal{D}$ includes weight values of the channels $j$, which are different from the weights of the channel $i$ (compared one-one respectively), $|\mathcal{D}|= d_{ij}$.
$\mathcal{A}^b_s$ and $\mathcal{A}^b_d$ are input activations for $\mathcal{S}$ and $\mathcal{D}$, respectively.
$P_j$ can be as 

\begin{equation} 
    \label{eq:output_1}
		P_j = \sum_{\mathcal{W}_s\in \mathcal{S}}\XNOR (\mathcal{A}^b_s,\mathcal{W}_s) + \sum_{\mathcal{W}_d\in \mathcal{D}}\XNOR (\mathcal{A}^b_d,\mathcal{W}_{d}).
\end{equation}

Because the input activation of the channel $i$ is the same as that of the channel $j$.
Suppose $\Bar{\mathcal{D}}$ includes weights of the channel $i$, which are different from that of the channel $j$, $|\mathcal{D}| = |\mathcal{\Bar{D}}| = d_{ij}$.
We can have $P_i$ as

\begin{equation} 
    \label{eq:output_1-1}
		P_{i} = \sum_{\mathcal{W}_s\in \mathcal{S}}\XNOR (\mathcal{A}^b_s,\mathcal{W}_s) + \sum_{\Bar{\mathcal{W}}_d\in \Bar{\mathcal{D}}}\XNOR (\mathcal{A}^b_d,\Bar{\mathcal{W}_{d}}),
\end{equation}

In consequence, $\sum_{\mathcal{W}_s\in \mathcal{S}}\XNOR (\mathcal{A}^b_s,\mathcal{W}_s)$ can be calculated as,
\begin{equation} 
    \label{eq:output_2}
		\sum_{\mathcal{W}_s\in \mathcal{S}}\XNOR (\mathcal{A}^b_s,\mathcal{W}_s) = P_i - \sum_{\Bar{\mathcal{W}}_d\in \Bar{\mathcal{D}}}\XNOR (\mathcal{A}^b_d,\Bar{\mathcal{W}_{d}}),
\end{equation}

and $\forall$ input activations, based on the characteristics of the XNOR operation, we have

\begin{equation} 
    \label{eq:output_3}
    \sum_{\mathcal{W}_d\in \mathcal{D}}\XNOR (\mathcal{A}^b_d,\mathcal{W}_{d}) + \sum_{\Bar{\mathcal{W}}_d\in \Bar{\mathcal{D}}}\XNOR (\mathcal{A}^b_d,\Bar{\mathcal{W}_{d}}) = d_{ij}.
\end{equation}
Use Eq. (\ref{eq:output_2}) and Eq. (\ref{eq:output_3}), we can reformulate the Eq. (\ref{eq:output_1}) as
\begin{equation} 
P_j = P_i - d_{ij} + 2\sum_{\mathcal{W}_d\in \mathcal{D}}\XNOR (\mathcal{A}^b_d,\mathcal{W}_{d}).
\end{equation}

In Sec. {\color{red}2}, we have $P_{ij} = \sum_{\mathcal{W}_d\in \mathcal{D}}\XNOR (\mathcal{A}^b_d,\mathcal{W}_{d})$.
Thus, we finally have the following equation.
\begin{equation} \tag{3}
    \label{eq:output_final}
		Y_j = 2(P_i - d_{ij} + 2P_{ij})-C_{in}\times M\times M.
\end{equation}
\textbf{Additional results}

\textbf{Effect of the number of centers.}
\begin{table}[] 
\begin{center}
\footnotesize
\begin{tabular}{lcccc}
\hline

\#centers&\multicolumn{1}{l}{MST-depth}      & \multicolumn{1}{l}{\begin{tabular}[c]{@{}c@{}}\#Params\\ (Mbit)\end{tabular}} & \multicolumn{1}{l}{\begin{tabular}[c]{@{}c@{}}\#Bit-Ops\\ (GOps)\end{tabular}} & \multicolumn{1}{c}{\begin{tabular}[c]{@{}c@{}}Top-1 Acc.\\ mean $\pm$ std (\%)\end{tabular}} \\ \hline
1      & $22.3$         & $0.545$             & $0.119$              & $91.49\pm 0.04$              \\ \hline
2      & $30.3$         & $0.550$             & $0.118$              & $91.45\pm 0.08$              \\ \hline
4      & $47.7$         & $0.574$             & $0.125$              & $91.42\pm 0.06$              \\ \hline
6      & $60.3$         & $0.581$             & $0.130$              & $91.53\pm 0.07$              \\ \hline
8      & $73.0$         & $0.607$             & $0.136$              & $91.49\pm 0.04$              \\ \hline

\end{tabular}
\end{center}
\caption{Accuracy, MST depth, number of parameters and bit-Ops \textit{w.r.t.} different number of centers on CIFAR-10 VGG-small model.}
\label{tab:comparison_center}
\end{table}
In this section, we provide an additional experimental results related to the effect of the number of initial centers for the training.
In particular, we do the training on VGG-small model and CIFAR-10 dataset with different number of centers, while $\lambda$ is fixed at 4e-6.
Besides, each number of centers, we execute the training three times and get the mean value for the report.

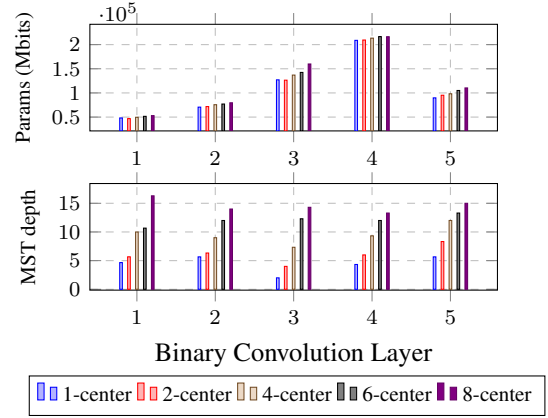
\begin{figure}[t] 
\begin{center}
        \pgfplotsset{compat=newest}
        \begin{tikzpicture}
        \begin{groupplot}[group style={group size=1 by 2,horizontal sep = 1.2cm, vertical sep = 0.7cm},height=3cm,width=7cm, ]
        \nextgroupplot[
        	x tick label style={/pgf/number format/1000 sep=},
        	ylabel=Params (Mbits),
        	enlargelimits=0.15, grid=both,
        	legend style={at={(0.5,-0.5)},
        		anchor=north,legend columns=-1},
                every tick label/.append style={font=\footnotesize},
                every axis legend/.append style={font=\footnotesize},
                ylabel style={font=\footnotesize},
        	ybar,
                grid style=dashed,
        	bar width=1pt,
        ]
        \addplot
        	coordinates {(1,48250) (2,70600)
        		 (3,127104) (4,208852) (5,89820)};
        \addplot
        	coordinates {(1,46827) (2,71768)
        		 (3,126416) (4,209543) (5,95229)};
        \addplot
        	coordinates {(1,49611) (2,75646)
        		 (3,136973) (4,213442) (5,98290)};
        \addplot
        	coordinates {(1,51553) (2,77019)
        		 (3,142464) (4,216600) (5,105077)};
        \addplot
        	coordinates {(1,52988) (2,79587)
        		 (3,159930) (4,216369) (5,110352)};
        
        \nextgroupplot[
        	x tick label style={/pgf/number format/1000 sep},
        	ylabel=MST depth,
        	enlargelimits=0.15, grid=both,
        	legend style={at={(0.5,-0.7)},anchor=north,legend columns=-1},
                every tick label/.append style={font=\footnotesize},
                every axis legend/.append style={font=\footnotesize},
        	ybar,
                grid style=dashed,
        	bar width=1pt,
                xlabel= Binary Convolution Layer,
                legend image post style={scale=0.8},
                every axis legend/.append style={font=\footnotesize,at={(-0.15,-0.8)}, anchor=north west,legend columns = 5},
                ylabel style={font=\footnotesize},
        ]
        \addplot
        	coordinates {(1,4.66) (2,5.66)
        		 (3,2) (4,4.33) (5,5.66)};
        \addplot
        	coordinates {(1,5.67) (2,6.33)
        		 (3,4) (4,6) (5,8.33)};

        \addplot
        	coordinates {(1,10) (2,9)
        		 (3,7.33) (4,9.33) (5,12)};
        \addplot
        	coordinates {(1,10.67) (2,12)
        		 (3,12.3) (4,12) (5,13.3)};
        \addplot
        	coordinates {(1,16.3) (2,14)
        		 (3,14.3) (4,13.3) (5,15)};

        \legend{1-center,2-center,4-center,6-center,8-center}
        \end{groupplot}
        \end{tikzpicture}
    
\end{center}
\caption{Number of parameters and MST depth on each convolution layer \textit{w.r.t.} different number of centers.}
\label{fig:center}
\end{figure}

Table \ref{tab:comparison_center} provides the MST depth, number of parameters, bit-ops and accuracy \textit{w.r.t.} different number of centers.
Accordingly, the MST depth, number of parameters and bit-ops tend to increase as the number of centers increases.
Specifically, when the number of centers changes from $1$ to $8$, the MST depth increases $3.27\times$, the number of parameters and bit-ops increase $1.11\times$ and $1.14\times$, respectively.
Meanwhile, accuracy barely changes with different number of centers.
For each binary convolution layer, as shown in Figure \ref{fig:center}, as the number of centers increases, both the MST depth and number of parameters also increase.
These findings suggest that opting for a single center is the most effective strategy to minimize MST depth, parameters, and bit-ops while preserving accuracy.

\end{document}